\documentclass[a4paper, 10 pt, conference]{hsmr} 
\usepackage[utf8]{inputenc}
\IEEEoverridecommandlockouts                       
\usepackage{mathtools}
\usepackage{graphicx}
\usepackage{float} 
\usepackage{booktabs}
\usepackage{subfigure}
\usepackage{geometry}
\usepackage[labelsep=space]{caption}
\usepackage{newtxmath,newtxtext}
\usepackage{authblk}
\usepackage{pgfplots}
\usepackage{todonotes}
\usepackage{graphics}
\usepackage{cite}
\usepackage{setspace}
\usepackage[font=footnotesize]{subfig}
\usepackage{mdwtab}
\usepackage{eqparbox}
\usepackage{fixltx2e}
\usepackage{stfloats}
\usepackage{color}
\usepackage{bbm}
\usepackage{multirow}
\usepackage{float}
\usepackage{caption}
\usepackage[normalem]{ulem}

\usepackage{newtxtext,newtxmath}

\pgfplotsset{compat=newest} 
 
\title{ \bf 
Deep Reinforcement Learning Based Semi-Autonomous Control for Robotic Surgery} 

\author{\LARGE Ruiqi Zhu$^{1}$}
\author{\LARGE Dandan Zhang$^{1,2}$} 
\author{\LARGE Benny Lo$^{1}$} 

\affil{ \textit{$^{1}$The Hamlyn Centre, Imperial College London}\\\textit{$^{2}$Department of Engineering Mathematics, University of Bristol}}

\begin{document}

\maketitle
\thispagestyle{empty}
\pagestyle{empty}

 

\section*{INTRODUCTION}



In recent year, autonomy has been widely introduced into surgical robotic systems to assist surgeons to carry out complex tasks reducing the workload during surgical operation \cite{yip2019robot}.
Most of the existing methods normally rely  on learning from demonstration \cite{chen2020supervised}, which requires a collection of Minimally Invasive Surgery (MIS) manoeuvres from expert surgeons. However, collecting such a dataset to regress a template trajectory can be tedious and may induce significant burdens to the expert surgeons.

In this paper, we propose a semi-autonomous control framework for robotic surgery and evaluate this framework in a simulated environment. We applied deep reinforcement learning methods to train an agent for autonomous control, which includes simple but repetitive manoeuvres. 
Compared to learning from demonstration, deep reinforcement learning can learn a new policy by altering the goal via modifying the reward function instead of collecting new dataset for a new goal. In addition to the autonomous control, we also created a handheld controller for manual precision control. The user can seamlessly switch to manual control at any time by moving the handheld controller. Finally, our method was evaluated in a customized simulated environment to demonstrate its efficiency compared to full manual control.

\section*{MATERIALS AND METHODS}



\begin{figure}
\centering
\includegraphics[width=1\columnwidth]{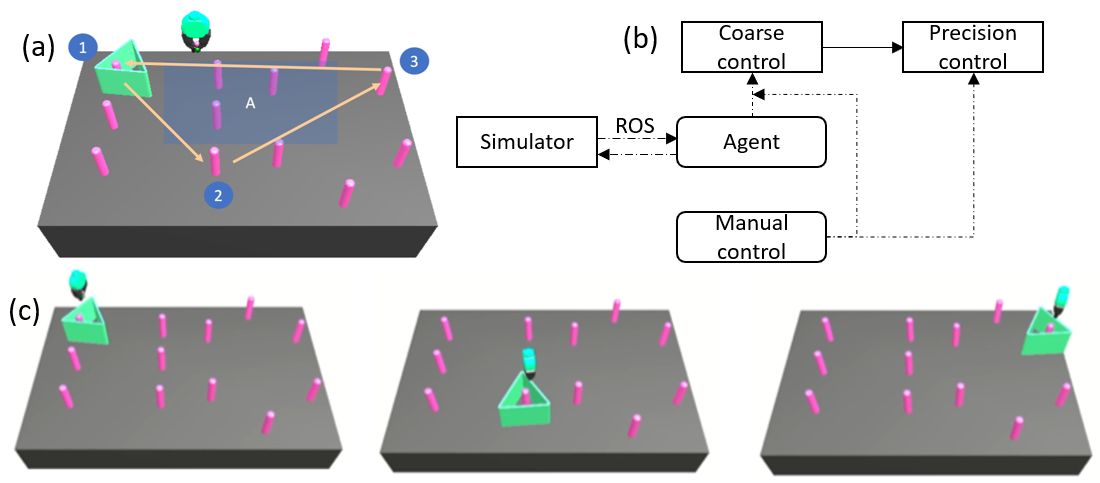}
\caption{The illustration of the evaluation task(a), the control flow chart(b), the final frame of episodes with different initialization (c).}
\label{task illustration}
\end{figure}

The customized simulator is developed based on Asynchronous Multi-Body Framework (AMBF) \cite{munawar2019real} as shown in Fig.\ref{task illustration} (a).  The aim is to implement semi-autonomous control for the peg transfer task. The task is segmented into two parts, automatic coarse control and manual override precision control. The coarse control includes controlling the gripper to approach the peg and modify its orientation to an appropriate pose for a grasp. The precision control includes fine-tuning the gripper's orientation, grasping and transferring the peg. The control flow chart is shown in Fig.\ref{task illustration} (b). For training an agent to operate in the simulator with deep reinforcement learning methods, we built an environment via Robot Operating System (ROS). With the interface, the environment can feedback reward, image frame and information telling whether the termination state is reached. Double Deep Q Network (DDQN) \cite{van2016deep} was used to optimize the agent for automating the coarse control. In addition, a handheld controller was developed for the user to override the system and carry out precision control.
 

\textbf{Coarse Control.}
For the coarse control, we considered it as a Markov Decision Process defined by a tuple \{$S,A,T,R,\gamma$\} which represents state space, action space, transition probability, reward function and discount factor. In this experiment, as it was visual-based, the agent only received an image frame after taking a step without knowing the actual state information. Visual perception offers the agent the potential of inferring the varying target state. In this experiment, we clipped the frame to the region of interest to reduce the computation load and then stacked four consecutive frames as the input to the deep neural network, so that it can infer the actual state.  We would like to hold the end-effector at a consistent height since we want to avoid the danger of the end-effector colliding with other objects. 
The action space was \{$dx,dy,d\phi$\}, the position movement along $x$ and $y$ axis in Cartesian space and the roll angle of the end-effector in Euler space. The action space was discretized with a precision of $6mm, 8mm, 10 rads$ respectively with ranges $[-6mm,6mm], [-8mm,8mm], [-10rads,10 rads]$. Narrowing the action space by discretization can bring faster convergence and save training time and computation. To encourage the agent to approach the target, and modify its orientation when the distance $d$ is less than the threshold $d_{threshold}$ of $10mm$
, the reward function was defined as shown in Equation \ref{reward function}, where $d_{t}$, $\triangle\theta_{t}$ refer to the distance to the target and the deviation to the desired orientation angle which is perpendicular to the closest side of the target at time step $t$ respectively. The discount factor $\gamma$ was set as 0.95.

\begin{equation}\label{reward function}
	r_{t+1} = \begin{cases}
	(d_{t}-d_{t+1})|d_{t}-d_{t+1}|, &\textsl{if $d_{t+1}>d_{threshold}$}\\
	(\triangle\theta_{t}-\triangle\theta_{t+1})|\triangle\theta_{t}-\triangle\theta_{t+1}|, &\textsl{otherwise}
		   \end{cases}
\end{equation}

DDQN was used to optimize the objective $J_{\theta}=\sum_{t=0}^{T-1}\gamma^{t}r_{t+1}$. The action value update equation is shown as following, where $a^{*}_{t+1}=\mathop{\arg\max_{{a_{t+1}}}} Q(s_{t+1},a_{t+1}|\theta)$. 

\begin{equation}\label{q function update}
    Q(s_{t},a_{t}|\theta)=r(s_{t},a_{t},s_{t+1})+ \gamma Q(s_{t+1},a^{*}_{t+1}|\theta') 
\end{equation}

The decoupling of the selection of the best action and the action value estimation of next state can reduce overestimation and therefore stabilize the training process. In addition, the use of target network $\theta'$ can further stabilize the training \cite{mnih2015human}.

\textbf{Precise Control.}
For the manual override precision control, we designed and developed a handheld controller as shown in Fig.\ref{camera_setting} (a). A depth camera was used to track the 3-D position of the tooltip of the 3-D printed handheld controller using library \textsl{OpenCv}, and an IMU sensor was attached at the end of the controller to track its 3-D orientation. Then, the pose was mapped onto the gripper in the simulator for the manual override control. In addition, a footpedal was used to control the clutch of the gripper.

\section*{RESULTS}

\begin{figure}
\centering
\includegraphics[width=1\columnwidth]{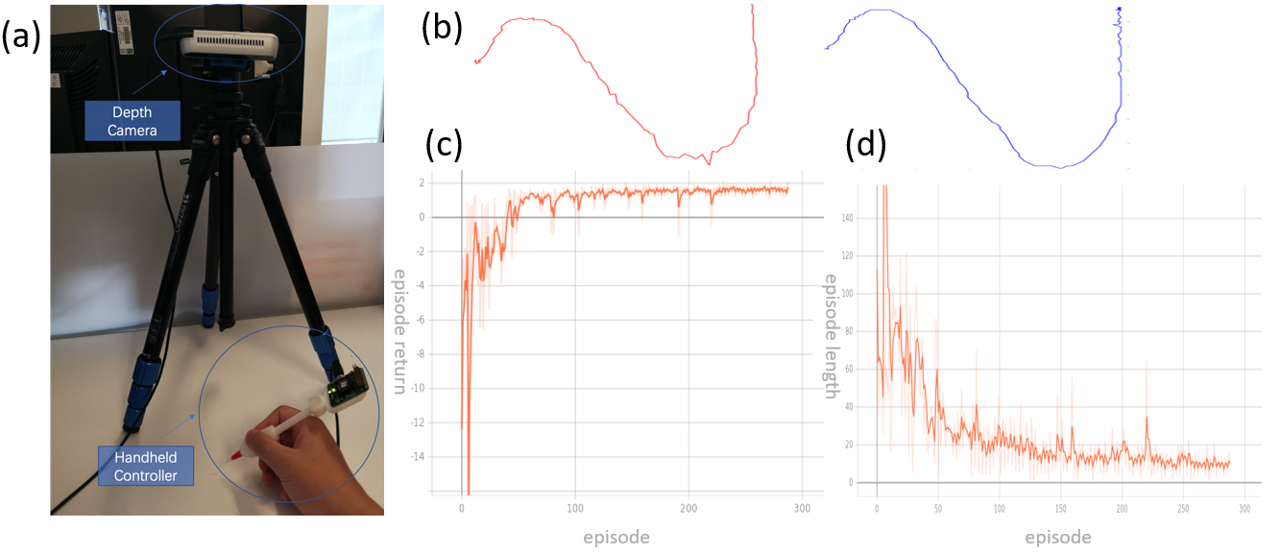}
\caption{The setup for manual override control (a), the qualitative results (b): gripper trajectory (red), controller trajectory (blue), episode vs. episode return (c), episode vs. episode length (d).}
\label{camera_setting}
\end{figure}

The training of the agent took around $150$ episodes to reach the convergence, as shown in Fig.\ref{camera_setting} (c). In addition, after the convergence, the steps required to complete an episode also converged indicating that it has learned a stable and efficient policy as shown in Fig.\ref{camera_setting} (d). The final frames of episodes with different initialized target positions are shown as Fig.\ref{task illustration} (c). For all three different target positions, the agent can successfully control the gripper to approach the target and modify its orientation to an appropriate position for a grasp. 

As for the manual override control, we evaluated the correspondence between the mapped gripper trajectory and the controller trajectory qualitatively as shown in Fig.\ref{camera_setting} (b). It indicates that the gripper trajectory can correspond to the controller trajectory. 

\begin{table}
    \renewcommand{\arraystretch}{1.2}
        \caption{Evaluation Results}
        \label{table:metrics}
    \centering
    \begin{tabular}{ccc}
    \toprule
    & \textbf{Manual}&\textbf{Semi-autonomous}\\
    \hline
    $M$ & 329mm & 136mm\\
    $T$ & 94s & 76s\\
    \bottomrule
    \end{tabular}
    \vspace{-1.5em}
\end{table}

We have conducted a user study to validate the proposed framework. The evaluation task is illustrated in Fig.\ref{task illustration} (a). First, the gripper needs to grasp the target at position $1$ and transfer the target to position $2$. After that, the gripper is reset to a position within region $A$. The process is repeated to transfer the target from position $2$ to position $3$ and from position $3$ to position $1$. Participants were asked to carry out this procedure for $9$ times. The average controller travel length $M$ and task completion time $T$ were recorded for evaluation. The evaluation results are shown on Table \ref{table:metrics}. It indicated that with the proposed framework, the travel length was reduced by around $58.7\%$ and the completion time was reduced by around $19.1\%$.

\section*{DISCUSSION}
In this paper, we proposed a deep reinforcement learning based semi-autonomous control framework. It uses the DDQN to implement the automatic coarse control while the user only need to focus on fine control and make the decision at critical points. The user study showed that the method can reduce the controller travel length by a great margin and the completion time as well. This demonstrates the potential of the proposed method in automating repetitive tasks and reducing the cognitive loads on the surgeons in  MIS operations. However, the reduction margin of the completion time was not as high as expected and this was because when starting the fine control after the coarse control phase, the user usually needed to identify the relative pose of the end-effector to the target by moving the controller slightly. Thus, future work includes enabling seamless collaborative control by offering visual or force feedback. In addition, further work will be carried out on transferring the learned policy to the da Vinci surgical Robotic platform. 

\nocite{*}
\bibliographystyle{IEEEtran}
\bibliography{references}

\end{document}